\documentclass[sigconf,natbib]{acmart}
\makeatletter
\renewcommand\@formatdoi[1]{\ignorespaces}
\makeatother

\AtBeginDocument{%
  \providecommand\BibTeX{{%
    \normalfont B\kern-0.5em{\scshape i\kern-0.25em b}\kern-0.8em\TeX}}}

\setcopyright{acmcopyright}
\copyrightyear{2019}
\acmYear{2019}

\acmConference[SIGIR 2019]{SIGIR '19: FACTS-IR Workshop}{July 25, 2019}{Paris, France}
\acmDOI{10.475/123_4}
\acmISBN{123-4567-24-567/08/06}

\usepackage{paralist}
\usepackage{enumitem}
\usepackage{multirow}
\usepackage{balance}
\usepackage{subfigure}
\usepackage[np, autolanguage]{numprint}
\usepackage{algorithm}
\usepackage{algorithmicx}

\usepackage{acronym}
\acrodef{GBDTs}{gradient boosting decision trees}
\acrodef{IR}{information retrieval}
\acrodef{MC-BRP}{Monte Carlo Bounds for Reasonable Predictions}
\newcommand{\MyTodo}[1]{\textcolor{red}{#1}}

\begin{document}

\title{Explaining Predictions from Tree-based Boosting Ensembles}

\author{Ana Lucic}
\orcid{}
\affiliation{%
  \institution{University of Amsterdam}
  \city{Amsterdam}
  \country{The Netherlands}
}
\email{a.lucic@uva.nl}

\author{Hinda Haned}
\orcid{}
\affiliation{%
  \institution{Ahold Delhaize}
  \city{Zaandam}
  \country{The Netherlands}
}
\email{hinda.haned@aholddelhaize.com}

\author{Maarten de Rijke}
\orcid{0000-0002-1086-0202}
\affiliation{%
  \institution{University of Amsterdam}
  \city{Amsterdam}
  \country{The Netherlands}
}
\email{derijke@uva.nl}


\begin{abstract}
Understanding how ``black-box'' models arrive at their predictions has sparked significant interest from both within and outside the AI community. Our work focuses on doing this by generating local explanations about individual predictions for tree-based ensembles, specifically Gradient Boosting Decision Trees (GBDTs). Given a correctly predicted instance in the training set, we wish to generate a counterfactual explanation for this instance, that is, the minimal perturbation of this instance such that the prediction flips to the opposite class. Most existing methods for counterfactual explanations are (1) model-agnostic, so they do not take into account the structure of the original model, and/or (2) involve building a surrogate model on top of the original model, which is not guaranteed to represent the original model accurately. There exists a method specifically for random forests; we wish to extend this method for GBDTs. This involves accounting for (1) the sequential dependency between trees and (2) training on the negative gradients instead of the original labels. 
\end{abstract}




\maketitle

\section{Introduction}
As \ac{IR} systems become more and more prevalent, it becomes increasingly important to understand how an \ac{IR} system produces a particular prediction and what exactly drives it to do so. 
Understanding how ``black-box'' models arrive at their predictions has sparked significant interest from both within and outside the \ac{IR} community. 
This can be in the context of rankings~\citep{terhoeve-2017-news}, recommendations~\citep{tintarev_explaining_2007}, or digital assistants that engage in interactive question answering~\citep{qu_answer_2019}.  

Explanations of an \ac{IR} system can be provided for the system as a whole or for individual decisions produced by the system. 
Explanations based on interpreting the model in all regions of the input space are called \emph{global explanations}, while those based on interpreting individual predictions are called \emph{local explanations}~\citep{guidotti-2018-survey}. 
The explainability problem is often cast in terms of supervised prediction models: IR systems usually involve a prediction at some point in the pipeline (i.e., predicting whether a document is relevant or not). 

Given how often we use complex models to help us make difficult decisions, it is important to be able to understand what happens during the training phase of the model. 
We propose doing this by generating local explanations about individual predictions. 
Recent work on local explanations is usually conducted in either a model-agnostic or model-specific way~\citep{guidotti-2018-survey}. 
Model-agnostic explanations typically involve approximating the original ``black-box'' model locally in the neighborhood of the instance in question~\citep{ribeiro-2016-should}, while model-specific explanations use the inner workings of the original ``black-box'' to explain the prediction of the given instance~\citep{tolomei-interpretable-2017}. 
The obvious advantage of model-agnostic explanations is that they can be applied to any type of model \citep{riberio-2016-model}, but since the explanation is based on a local approximation of the original model, there exists some inherent degree of error between the original model and the local approximation. 
Indeed, since the local model is an approximation, there is no guarantee that it is appropriately representative of the original model, especially in other parts of the input space~\citep{alvarez-melis_robustness_2018}. 
In our work, we focus on generating model-specific explanations for boosting ensembles since they are widely used in industry and have demonstrated superior performance in a wide range of tasks. 

This gives rise to our leading research question:
\begin{quotation}
\em
How can we automatically generate actionable explanations for individual predictions of tree-based boosting ensembles?
\end{quotation}

\section{Agenda}
In order to address our leading research question, we propose a three-part research agenda for using explanations to understand individual predictions. 
\begin{enumerate}
	\item Generate explanations for individual ``black-box'' predictions in terms of
		\begin{inparaenum}[(i)]
			\item why a particular prediction was classified as a certain class, 
			\item what it would have taken for the prediction to be classified as the alternative class, and
			\item how to perturb the model in order to change the prediction. 
		\end{inparaenum}
	\item Develop a mechanism that allows the user to change the prediction based on the explanation. 
	\item Evaluate the effectiveness of such explanations on users' confidence in and trust of the original ``black-box'' model. This also involves determining appropriate baselines and metrics, and a sensible experimental environment in terms of the people involved and the questions asked. 
\end{enumerate}
In this work, we outline ideas along with a case study about items (1) and (2) above. 
The work of \citet{tolomei-interpretable-2017} has the potential to solve this problem but we argue that it
\begin{inparaenum}[(i)]
	\item does not apply to boosting ensemble methods, and
	\item has scalability issues. 
\end{inparaenum}
In order to come up with a satisfactory solution to our problem, we take the method from~\citep{tolomei-interpretable-2017}, explain it and articulate how it could be extended to accommodate tree-based boosting ensembles. 
In this extended abstract, we focus on adaptive boosting \citep{hastie_multi-class_2009} first to disentangle the sequential training nature of boosting methods. 

\section{A Case Study in Explaining Individual Predictions -- Work in Progress}
We focus on explaining predictions from tree-based boosting ensemble methods (or simply boosting methods). 
Boosting methods are based on sequentially training (weak) models that, in each iteration, focus more on correcting the mistakes of the previous model. 
We train a boosting ensemble $\hat{f}$ using an input set $X$ to predict a target variable $y \in \left\{-1, 1\right\}$, where $\left\{T_1, \dots, T_K\right\}$ are the set of base classifiers for the ensemble and $\left\{\hat{h}_1, \dots, \hat{h}_K\right\}$ are the corresponding predictions for each base classifier. 

In adaptive boosting \citep{hastie_multi-class_2009}, each iteration $\hat{h}_k$ improves over $\hat{h}_{k-1}$ by upweighting misclassified instances (and downweighting correctly classified instances) by a factor of $e^{\alpha_k}$, where $\alpha_k$ is the weight assigned to $\hat{h}_k$ in the ensemble and is defined as
\begin{equation}
\label{equation:upweighting}
 \alpha_k = \log\frac{1 - err_k}{err_k}  + \log(K - 1)           
\end{equation}
and $err_k$ is the classification error of the $k$-th base classifier $h_k$.

\subsection{Problem definition}
\citet{tolomei-interpretable-2017} investigate the interpretability of random forests (RFs) by determining what drives a model to produce a certain output for a given instance in a binary classification task. 
They frame the problem in terms of actionable recommendations for transforming negatively labeled instances into positively labeled ones in a binary classification task. 
Our objective is to extend this method to work for boosting methods and later use these explanations to transform misclassified instances into correctly classified ones. 
This involves accounting for some components of boosting that do not apply to RFs:
\begin{inparaenum}[(i)]
	\item the sequential dependency between trees, and 
	\item training on the negative gradients instead of the original labels (in the case of \ac{GBDTs}). 
\end{inparaenum}
We break the task up into two stages:
\begin{enumerate}
	\item We first extend \citep{tolomei-interpretable-2017} to work for adaptive boosting \citep{hastie_multi-class_2009}, which still trains on the original labels. This allows us to focus specifically on training trees in sequence and use this to narrow our search space. 
	\item We extend our new method for adaptive boosting to gradient boosting~\citep{friedman_additive_2000}, where we not only train in sequence but also train on the negative gradients of the previous tree. 
\end{enumerate}
\noindent%
This leads to the following research questions:

\begin{itemize}
	\item \textbf{RQ1:} Given an instance, how can we perturb the instance such that the prediction for this instance flips from one class to another?
	\item \textbf{RQ2:} Given an instance, how can we perturb the model such that the prediction for this instance flips from one class to another?	

\end{itemize}

\subsection{Related work}
The method in \citet{tolomei-interpretable-2017} is defined as follows: let $x$ be an observation in the set $X$ such that $x$ is a true negative instance (i.e., $\hat{f}(x) = f(x) = -1$, where $\hat{f}$ is the overall prediction of the ensemble and $f$ is the true label). 
The objective is to create a new instance, $x'$, that is an $\epsilon$-transformation from an existing positively predicted instance, ${x}^{+}$. 

\begin{enumerate}
	\item The trees in the ensemble $\mathcal{T} = \{T_1, \dots, T_K\}$ (an RF in this case) can be partitioned into two sets depending on whether the prediction resulting from each tree $T_k$ is either positive or negative (the base classifier $\hat{h}_k$ corresponding to tree $T_k$ is either $+1$ or $-1$). 
We are interested in the set of trees that result in negative predictions since we want to determine the criteria for turning these into positive predictions. 
	\item Therefore, for every positive path $p_{k,j}^{+}$ (i.e., paths that result in a positive prediction, indexed by $j$) in every negative tree $T_k$ (i.e., $\hat{h}_k(x) = -1$), we want to generate an instance ${x^{+}_{j(\epsilon)}}$ that satisfies this positive path (i.e., $\hat{h}_k({x}^{+}_j) = +1$), based on our original instance $x$.
	\item We create ${x^{+}_{j(\epsilon)}}$ by examining the feature values of $x$ and the corresponding splitting thresholds in $p_{k,j}^{+}$. For each feature in $p_{k,j}^{+}$, if $x$ satisfies the splitting threshold for $p_{k,j}^{+}$, then we leave the value for this feature alone. If not, then we tweak the value for this feature such that it is $\epsilon$-away from the splitting threshold and satisfies $p_{k,j}^{+}$. 
	\item We construct an ${x^{+}_{j(\epsilon)}}$ based on $x$ from every positive path $j$ in every negative tree $k$ and evaluate the output of the entire ensemble using this ${x^{+}_{j(\epsilon)}}$. If $\hat{f}({x_{j(\epsilon)}})^+ = +1$ then ${x^{+}_{j(\epsilon)}}$ is a candidate transformation of $x$.
	\item We greedily choose the candidate transformation that is closest to the original instance and this is returned as the minimal perturbation of the original instance such that the prediction flips from negative to positive. We call this $x'$. 
	\item Since this $\epsilon$-perturbation allows us to discriminate between the two classes so it can be viewed as the contrastive explanation \citep{miller-2017-explanations} for why $\hat{f}(x) = -1$ as opposed to $+1$. 
\end{enumerate}
This work relies heavily on being able to enumerate the positive paths $p_{k,j}^{+}$ in each negative tree $T_k$, which is not possible when training on the negative gradients instead of the original labels. 
This is also very computationally intensive since we compute an $\epsilon$-transformation for an ${x}^{+}_j$ in each $p_{k,j}^{+}$. 
In our work, we want to use the sequential training nature of boosting methods to narrow the search space as early as possible. 

\subsection{Method outline}
Given an instance $x$, we are interested in reducing the search space for ${x}^{+}_j$ in order to make the method by \citet{tolomei-interpretable-2017} more scalable.
To this end, we look for a subset of the original ensemble, $\mathcal{T}\subseteq \left\{T_1, \dots, T_K\right\}$, such that the rest of the ensemble can safely be ignored. 
That is, we want to select the most important trees in the overall model without omitting trees that were particularly influential for this prediction.

\begin{figure}[h]
\centering
\includegraphics[scale=0.55]{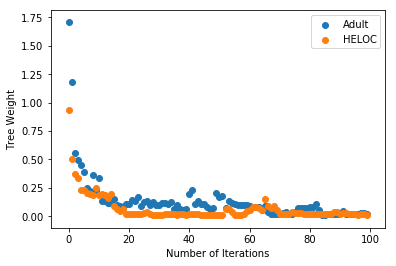}
\caption{The distribution of weights $\alpha_k$ for each iteration (or tree) in the ensemble.}
\label{fig:weights}
\end{figure}

We pursue two directions to determine whether such a $\mathcal{T}$ might be found. 
The first idea is to consider how much each tree contributes to the prediction by examining their corresponding weights $\left\{\alpha_1, \dots, \alpha_K\right\}$. 
We want to determine whether or not they decrease for each iteration $k$ in the training of the ensemble, and if so, how quickly this happens. 
The hypothesis is that if the weights drop quickly and to small quantities, then we can narrow the search space by only examining the trees in the beginning of the ensemble.
We choose two binary classification datasets: \emph{Adult}~\citep{adult-data} and \emph{home equity line of credit} (HELOC) \citep{heloc} and train an adaptive boosting model with 100 iterations, each with maximum depth 4 on the two datasets. 
Figure~\ref{fig:weights} shows the weights $\left\{\alpha_1, \dots, \alpha_K\right\}$ for each iteration in the model. 
Indeed, we see that the trees at the beginning of the ensemble seem to be more important to the overall prediction, as they have higher weights, than the trees towards the end. 
Therefore, if we want to reduce the search space, a sensible starting point would be to identify $K' < K$ based on the distribution of $\left\{\alpha_1, \dots, \alpha_K\right\}$ and examine only the first $K'$ trees in the ensemble. 
The potential error resulting from only considering the first $K'$ trees is sufficiently small given that the weights of the remaining trees are small, and therefore their impact on the overall prediction is minimal in comparison to the first $K'$ trees. 
In addition to giving us a way to reduce the search space, this $K'$ can also provide some insight into how difficult it was for the model to classify this instance -- the larger the $K'$, the more difficult it was. 
%

\begin{figure}[h!]
\begin{subfigure}
  \centering
  \includegraphics[scale=0.55]{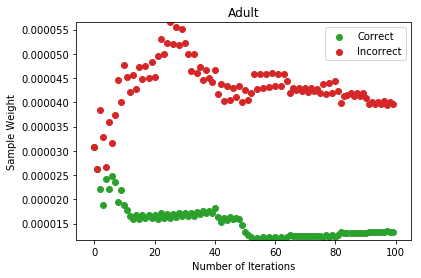}  
\end{subfigure}

\begin{subfigure}
  \centering
  \includegraphics[scale=0.5]{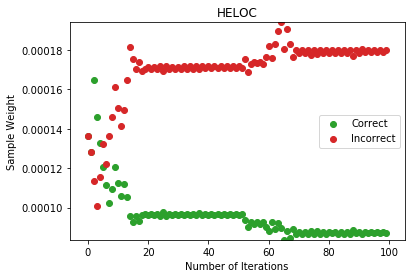}  
\end{subfigure}
\caption{The evolution of sample weights $w_k(x)$ for one correctly classified instance and one incorrectly classified instance in the Adult (above) and HELOC (below) datasets.}
\label{fig:sample-weights}
\end{figure}

Another option for determining the subset of trees $\mathcal{T}$ that would allow us to reduce the search space is by looking for structure in how the sample weights $\left\{w_1(x), \dots, w_K(x)\right\}$ change as an instance $x$ goes through each iteration $k$ of the model and identifying trees of interest based on this distribution. 
If the prediction of iteration $k$, $\hat{h}_k(x)$, is correct, then $w_k(x) < w_{k-1}(x)$; the opposite is true if $\hat{h}_k(x)$ is incorrect. 
Figure~\ref{fig:sample-weights} shows the evolution of these sample weights for two random instances in each of the datasets, \emph{Adult} and \emph{HELOC}: one that is correctly classified (depicted in green) and one that is incorrectly classified (depicted in red). 
We see that in both datasets, the weights for the correct instances decrease substantially within the first 15 iterations, implying the model is continuously classifying the instances correctly. 
In contrast, the weights for the incorrect instances increase substantially within this same period, implying the model is continuously misclassifying these instances. 
When the weights flatten out (e.g., for the correct instance in the \emph{Adult} dataset between $k = 18$ and $k = 40$), this implies $\hat{h}_k(x)$ is oscillating between $+1$ and $-1$, or analogously, oscillating between being correct and incorrect. 
The structure in the weight evolution of a particular instance gives us some insight into how the model learns to classify this point and how the the prediction fluctuates with each iteration. 
This can help us determine which trees should be included in the subset of the original ensemble we want to examine further and we outline some further ideas for this in Section 3.4. 

\subsection{Next steps}
We have provided some initial ideas for generating explanations for tree-based boosting predictions. 
We plan to investigate learning $K'$ for a given instance $x$, perhaps based on the distribution of training sample weights $\left\{w_k(x)\right\}_{k=1}^{K}$ along with the weights of each iteration $\left\{\alpha_k\right\}_{k=1}^{K}$. 
We also plan to investigate how this method could be extended to account for training on the negative gradients as is done in GBDTs. 

Finally, we plan to evaluate our method, and, in particular, the degree to which our proposed explanations are actionable, through a user study with participants from the MSc Data Science and MSc Artificial Intelligence programs at the University of Amsterdam.

\section{Conclusion}
We have sketched a research agenda for explaining predictions from boosting methods and sketched a case study to illustrate how to generate these explanations. 

In our case study, we examined how we can use the sequential training nature of boosting methods to narrow the search space for alternative examples when generating explanations. 
We will also explore how training on the negative gradient can be used to generate explanations for GBDT predictions and will evaluate the impact these types of explanations have on users who interact with the system. 
Finally, we invite the community to join the discussion on how we can automatically and transparently fix algorithmic errors, in ways that are meaningful for IR system experts as well as those outside the community.
\balance

\begin{acks}
This research was partially supported by
Ahold Delhaize,
the Association of Universities in the Netherlands (VSNU),
the Innovation Center for Artificial Intelligence (ICAI),
and
the Netherlands Organisation for Scientific Research (NWO)
under pro\-ject nr
652.\-001.\-003. 
All content represents the opinion of the authors, which is not necessarily shared or endorsed by their respective employers and/or sponsors.
\end{acks}

\bibliographystyle{ACM-Reference-Format}
\bibliography{ana-facts-ir}


\begin{thebibliography}{13}


\ifx \showCODEN    \undefined \def \showCODEN     #1{\unskip}     \fi
\ifx \showDOI      \undefined \def \showDOI       #1{#1}\fi
\ifx \showISBNx    \undefined \def \showISBNx     #1{\unskip}     \fi
\ifx \showISBNxiii \undefined \def \showISBNxiii  #1{\unskip}     \fi
\ifx \showISSN     \undefined \def \showISSN      #1{\unskip}     \fi
\ifx \showLCCN     \undefined \def \showLCCN      #1{\unskip}     \fi
\ifx \shownote     \undefined \def \shownote      #1{#1}          \fi
\ifx \showarticletitle \undefined \def \showarticletitle #1{#1}   \fi
\ifx \showURL      \undefined \def \showURL       {\relax}        \fi
\providecommand\bibfield[2]{#2}
\providecommand\bibinfo[2]{#2}
\providecommand\natexlab[1]{#1}
\providecommand\showeprint[2][]{arXiv:#2}

\bibitem[\protect\citeauthoryear{??}{adu}{1996}]%
        {adult-data}
 \bibinfo{year}{1996}\natexlab{}.
\newblock \bibinfo{title}{UCI Machine Learning Repository}.
\newblock   (\bibinfo{year}{1996}).
\newblock
\showURL{%
\url{https://archive.ics.uci.edu/ml/datasets/Adult}}


\bibitem[\protect\citeauthoryear{Alvarez-Melis and Jaakkola}{Alvarez-Melis and
  Jaakkola}{2018}]%
        {alvarez-melis_robustness_2018}
\bibfield{author}{\bibinfo{person}{David Alvarez-Melis} {and}
  \bibinfo{person}{Tommi~S. Jaakkola}.} \bibinfo{year}{2018}\natexlab{}.
\newblock \showarticletitle{On the {Robustness} of {Interpretability}
  {Methods}}.
\newblock \bibinfo{journal}{{\em arXiv:1806.08049 [cs, stat]\/}}
  (\bibinfo{date}{June} \bibinfo{year}{2018}).
\newblock
\newblock
\shownote{arXiv: 1806.08049.}


\bibitem[\protect\citeauthoryear{FICO}{FICO}{[n. d.]}]%
        {heloc}
\bibfield{author}{\bibinfo{person}{FICO}.} \bibinfo{year}{[n. d.]}\natexlab{}.
\newblock \bibinfo{title}{{Explainable} {Machine} {Learning} {Challenge}}.
\newblock   (\bibinfo{year}{[n. d.]}).
\newblock
\showURL{%
\url{https://community.fico.com/s/explainable-machine-learning-challenge?tabset-3158a=2}}


\bibitem[\protect\citeauthoryear{Friedman, Hastie, and Tibshirani}{Friedman
  et~al\mbox{.}}{2000}]%
        {friedman_additive_2000}
\bibfield{author}{\bibinfo{person}{Jerome Friedman}, \bibinfo{person}{Trevor
  Hastie}, {and} \bibinfo{person}{Robert Tibshirani}.}
  \bibinfo{year}{2000}\natexlab{}.
\newblock \showarticletitle{{Additive} {Logistic} {Regression}: {A}
  {Statistical} {View} of {Boosting}}.
\newblock  (\bibinfo{year}{2000}), \bibinfo{pages}{71}.
\newblock


\bibitem[\protect\citeauthoryear{Guidotti, Monreale, Turini, Pedreschi, and
  Giannotti}{Guidotti et~al\mbox{.}}{2018}]%
        {guidotti-2018-survey}
\bibfield{author}{\bibinfo{person}{Riccardo Guidotti}, \bibinfo{person}{Anna
  Monreale}, \bibinfo{person}{Franco Turini}, \bibinfo{person}{Dino Pedreschi},
  {and} \bibinfo{person}{Fosca Giannotti}.} \bibinfo{year}{2018}\natexlab{}.
\newblock \showarticletitle{A survey of methods for explaining black box
  models}.
\newblock \bibinfo{journal}{{\em arXiv preprint arXiv:1802.01933\/}}
  (\bibinfo{year}{2018}).
\newblock


\bibitem[\protect\citeauthoryear{Hastie, Rosset, Zhu, and Zou}{Hastie
  et~al\mbox{.}}{2009}]%
        {hastie_multi-class_2009}
\bibfield{author}{\bibinfo{person}{Trevor Hastie}, \bibinfo{person}{Saharon
  Rosset}, \bibinfo{person}{Ji Zhu}, {and} \bibinfo{person}{Hui Zou}.}
  \bibinfo{year}{2009}\natexlab{}.
\newblock \showarticletitle{Multi-class {AdaBoost}}.
\newblock \bibinfo{journal}{{\em Statistics and Its Interface\/}}
  \bibinfo{volume}{2}, \bibinfo{number}{3} (\bibinfo{year}{2009}),
  \bibinfo{pages}{349--360}.
\newblock
\showISSN{19387989, 19387997}


\bibitem[\protect\citeauthoryear{Miller}{Miller}{2019}]%
        {miller-2017-explanations}
\bibfield{author}{\bibinfo{person}{Tim Miller}.}
  \bibinfo{year}{2019}\natexlab{}.
\newblock \showarticletitle{Explanation in artificial intelligence: Insights
  from the social sciences}.
\newblock \bibinfo{journal}{{\em Artifical Intelligence\/}}
  \bibinfo{volume}{267} (\bibinfo{date}{February} \bibinfo{year}{2019}),
  \bibinfo{pages}{1--38}.
\newblock


\bibitem[\protect\citeauthoryear{Qu, Yang, Croft, Scholer, and Zhang}{Qu
  et~al\mbox{.}}{2019}]%
        {qu_answer_2019}
\bibfield{author}{\bibinfo{person}{Chen Qu}, \bibinfo{person}{Liu Yang},
  \bibinfo{person}{Bruce Croft}, \bibinfo{person}{Falk Scholer}, {and}
  \bibinfo{person}{Yongfeng Zhang}.} \bibinfo{year}{2019}\natexlab{}.
\newblock \showarticletitle{Answer {Interaction} in {Non}-factoid {Question}
  {Answering} {Systems}}.
\newblock \bibinfo{journal}{{\em Proceedings of the 2019 Conference on Human
  Information Interaction and Retrieval - CHIIR '19\/}} (\bibinfo{year}{2019}),
  \bibinfo{pages}{249--253}.
\newblock
\showDOI{%
\url{https://doi.org/10.1145/3295750.3298946}}
\newblock
\shownote{arXiv: 1901.03491.}


\bibitem[\protect\citeauthoryear{Ribeiro, Singh, and Guestrin}{Ribeiro
  et~al\mbox{.}}{2016a}]%
        {riberio-2016-model}
\bibfield{author}{\bibinfo{person}{Marco~Tulio Ribeiro},
  \bibinfo{person}{Sameer Singh}, {and} \bibinfo{person}{Carlos Guestrin}.}
  \bibinfo{year}{2016}\natexlab{a}.
\newblock \showarticletitle{Model-Agnostic Interpretability of Machine
  Learning}.
\newblock \bibinfo{journal}{{\em ICML Workshop on Human Interpretability in
  Machine Learning\/}} (\bibinfo{year}{2016}).
\newblock


\bibitem[\protect\citeauthoryear{Ribeiro, Singh, and Guestrin}{Ribeiro
  et~al\mbox{.}}{2016b}]%
        {ribeiro-2016-should}
\bibfield{author}{\bibinfo{person}{Marco~Tulio Ribeiro},
  \bibinfo{person}{Sameer Singh}, {and} \bibinfo{person}{Carlos Guestrin}.}
  \bibinfo{year}{2016}\natexlab{b}.
\newblock \showarticletitle{Why should {I} trust you?: {Explaining} the
  predictions of any classifier}. In \bibinfo{booktitle}{{\em KDD}}. ACM,
  \bibinfo{pages}{1135--1144}.
\newblock


\bibitem[\protect\citeauthoryear{ter Hoeve, Heruer, Odijk, Schuth, Spitters,
  and de~Rijke}{ter Hoeve et~al\mbox{.}}{2017}]%
        {terhoeve-2017-news}
\bibfield{author}{\bibinfo{person}{Maartje ter Hoeve}, \bibinfo{person}{Mathieu
  Heruer}, \bibinfo{person}{Daan Odijk}, \bibinfo{person}{Anne Schuth},
  \bibinfo{person}{Martijn Spitters}, {and} \bibinfo{person}{Maarten de
  Rijke}.} \bibinfo{year}{2017}\natexlab{}.
\newblock \showarticletitle{Do news consumers want explanations for
  personalized news rankings?}. In \bibinfo{booktitle}{{\em FATREC Workshop on
  Responsible Recommendation}}.
\newblock


\bibitem[\protect\citeauthoryear{Tintarev}{Tintarev}{2007}]%
        {tintarev_explaining_2007}
\bibfield{author}{\bibinfo{person}{Nava Tintarev}.}
  \bibinfo{year}{2007}\natexlab{}.
\newblock \showarticletitle{Explaining {Recommendations}}.
\newblock In \bibinfo{booktitle}{{\em User {Modeling} 2007}},
  \bibfield{editor}{\bibinfo{person}{Cristina Conati},
  \bibinfo{person}{Kathleen McCoy}, {and} \bibinfo{person}{Georgios Paliouras}}
  (Eds.). Vol.~\bibinfo{volume}{4511}. \bibinfo{publisher}{Springer Berlin
  Heidelberg}, \bibinfo{address}{Berlin, Heidelberg},
  \bibinfo{pages}{470--474}.
\newblock
\showISBNx{978-3-540-73077-4 978-3-540-73078-1}


\bibitem[\protect\citeauthoryear{Tolomei, Silvestri, Haines, and
  Lalmas}{Tolomei et~al\mbox{.}}{2017}]%
        {tolomei-interpretable-2017}
\bibfield{author}{\bibinfo{person}{Gabriele Tolomei}, \bibinfo{person}{Fabrizio
  Silvestri}, \bibinfo{person}{Andrew Haines}, {and} \bibinfo{person}{Mounia
  Lalmas}.} \bibinfo{year}{2017}\natexlab{}.
\newblock \showarticletitle{Interpretable {Predictions} of {Tree}-based
  {Ensembles} via {Actionable} {Feature} {Tweaking}}.
\newblock \bibinfo{journal}{{\em Proceedings of the 23rd ACM SIGKDD
  International Conference on Knowledge Discovery and Data Mining - KDD '17\/}}
  (\bibinfo{year}{2017}), \bibinfo{pages}{465--474}.
\newblock


\end{thebibliography}

\end{document}